# MATA (మాట): Mindful Assessment of the Telugu Abilities of Large Language Models

**Chalamalasetti Kranti[1], Sowmya Vajjala[2]**

[1]Department of Linguistics, University of Potsdam, Germany,

[2]National Research Council, Ottawa, Canada

kranti.chalamalasetti@uni-potsdam.de, sowmya.vajjala@nrc-cnrc.gc.ca

## Abstract

In this paper, we introduce MATA, a novel evaluation dataset to assess the ability of Large Language Models (LLMs) in Telugu language, comprising $729$ carefully curated multiple-choice and open-ended questions that span diverse linguistic dimensions. We evaluate 11 open-weight and closed-source LLMs on our dataset and present a fine-grained analysis of their performance. Further, we empirically show how LLMs rely on superficial heuristics such as answer position and distractor patterns for multiple-choice questions. Finally, we also compare LLM-as-a-judge evaluation with human evaluation for open-ended questions assess its reliability in a low-resource language. We argue that such fine-grained evaluation is essential for understanding model limitations and can inform the development of more linguistically capable LLMs, while also serving as a foundation for future research in Telugu NLP. Our dataset is available at: https://huggingface.co/datasets/TeluguLLMResearch/MATA

## 1. Introduction

The rapid rise of LLMs has led to a continual stream of new models, each claiming state-of-the-art performance on standard benchmark datasets. While most of the datasets are in English, several dataset creation efforts, such as translated versions of MMLU (Hendrycks et al., 2021; Singh et al., 2024b) aim to develop multilingual benchmarks across languages. However, the reliability and comprehensiveness of such large-scale benchmarks warrant closer scrutiny, as shown by some recent work (Liang et al., 2023; Balloccu et al., 2024; Alzahrani et al., 2024). Other research has identified significant shortcomings in some standard datasets including issues like time-sensitive or factually incorrect questions and misaligned answers, which may lead to unjust penalization of models (Gema et al., 2025; Plaza et al., 2024; Cengiz et al., 2025; Kranti et al., 2025). Moreover, translations, regardless of their linguistic fidelity, often fail (Naous et al., 2024) to capture the cultural and linguistic nuances specific to these languages. To address such limitations, several recent efforts (Alghallabi et al., 2025; Sibaee et al., 2025; Alghamdi et al., 2025; Tran et al., 2025) have focused on creating native test sets for non-English languages. These datasets aim to incorporate culturally and linguistically grounded elements (Cheng et al., 2025; Adilazuarda et al., 2025), thereby offering more representative benchmarks for evaluating multilingual LLMs. Such initiatives (Liu et al., 2025; Susanto et al., 2025) emphasize the need for evaluation resources that reflect the unique linguistic features and cultural context of each language.

Following this lead, we focus on the evaluation of LLMs in Telugu in this paper. Telugu, a Dravidian language, is the 18th most spoken language in the world (Ethnologue, 2025), but it is typically considered low-resource in terms of NLP support. It exhibits complex grammar (e.g., different forms of word compounding), and culturally embedded reading comprehension formats (e.g., metrical poetry) which can be challenging to LLMs. Existing (non-translated) multilingual benchmarks that include Telugu focus primarily on multiple-choice questions that assess factual knowledge (Romanou et al., 2025; Verma et al., 2024) without evaluating any of the broader linguistic abilities, or other culturally grounded aspects such as riddles, idioms and proverbs, leading to an incomplete assessment of capabilities in the language.

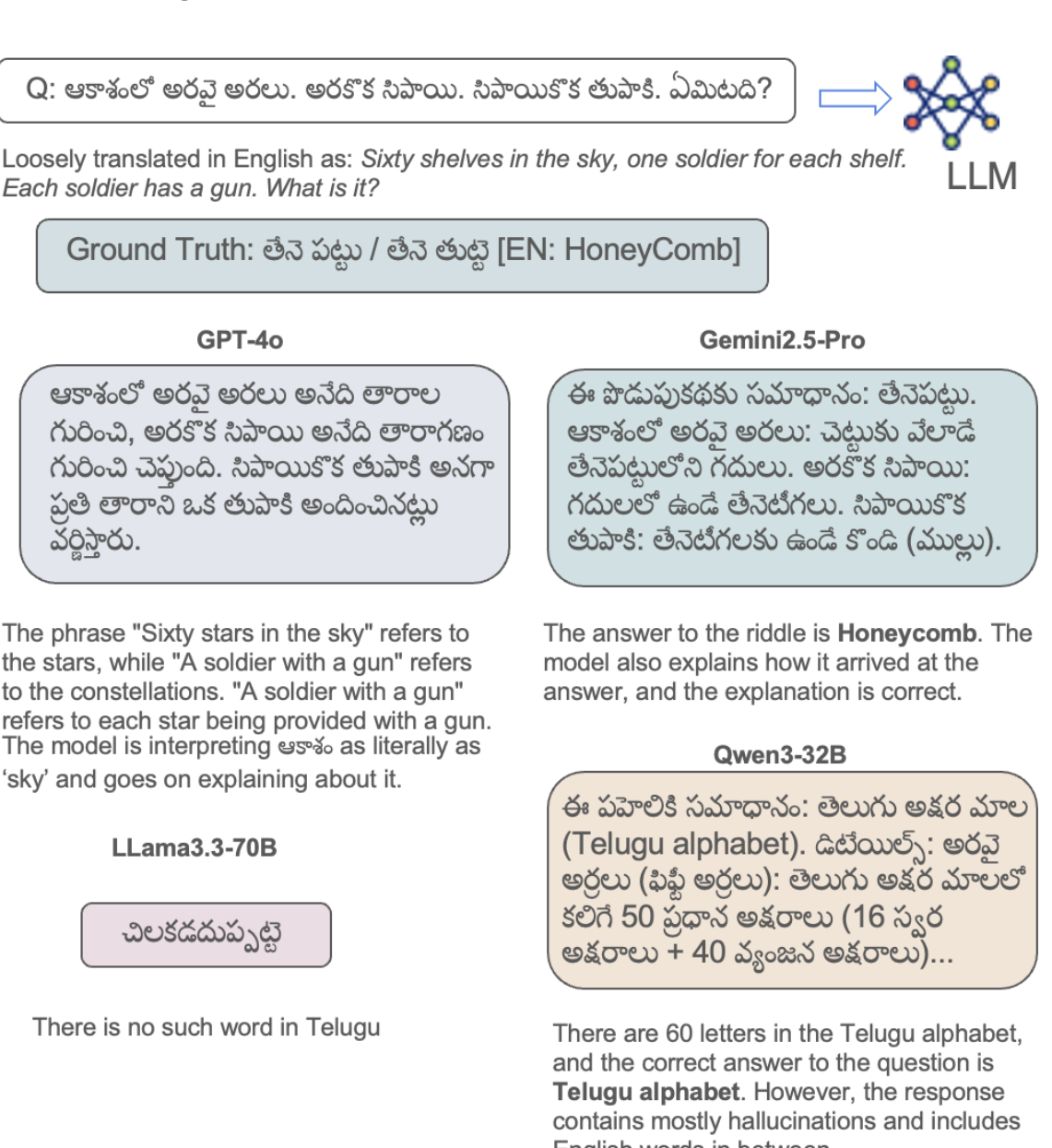

Figure 1: An open-ended Telugu riddle question from MATA, sourced from a 1st Grade Telugu text book, along with a few LLM responses, and comments (in English) about the output.

To address these limitations, we introduce MATA (మాట)[1], a new evaluation dataset specifically designed to assess a wide spectrum of linguistic competencies in Telugu. Figure 1 shows an example question from MATA, sourced from a 1st Grade Telugu textbook, and the responses from different LLMs, indicating how Telugu questions considered appropriate in complexity for a first grader can still be challenging for today's LLMs.

**Contributions:** Our contributions are threefold. First, we present a new evaluation dataset that systematically targets various dimensions of linguistic understanding in Telugu for the first time. Second, we systematically benchmark 11 proprietary and open-weight LLMs on this dataset. Third, we study the robustness of the multiple-choice question format and the reliability of LLM-as-a-judge to evaluate open-ended questions in a low-resource language context.

## 2. Related Work

Recent efforts in multilingual LLM evaluation have emphasized the importance of culturally and linguistically grounded datasets. Several studies have proposed improved evaluation resources for Arabic, focusing on aspects such as poetry (Alghallabi et al., 2025), cultural context (Sibaee et al., 2025), and regional expressions (Alghamdi et al., 2025). Similar efforts have been made for Italian (Magnini et al., 2025), Korean (Shin et al., 2025), Polish (Dadas et al., 2025), Bengali (Kabir et al., 2025), Tibetan (Gao et al., 2025) and Irish (Tran et al., 2025) with datasets designed to capture linguistic phenomena unique to the language, including idiomatic usage and syntactic patterns. In the case of Chinese (Huang et al., 2023; Zhao et al., 2025), curated benchmarks have explored classical language constructs, idioms, and culturally significant knowledge that are often missed in translated benchmarks (Agarwal et al., 2025).

Our work extends these lines of research by focusing on Telugu. Unlike prior datasets involving Telugu (Endait et al., 2025) that either rely on translations (Singh et al., 2024b) or are automatically crawled from web sources (Romanou et al., 2024; Singh et al., 2024a), we construct a test suite that directly engages with Telugu-specific grammatical structures, semantic patterns, and culturally embedded language use by sourcing our questions from school grade language textbooks and higher level exams. To the best of our knowledge, MATA is the first evaluation dataset to offer a nuanced, multifaceted assessment of LLM capabilities in Telugu.

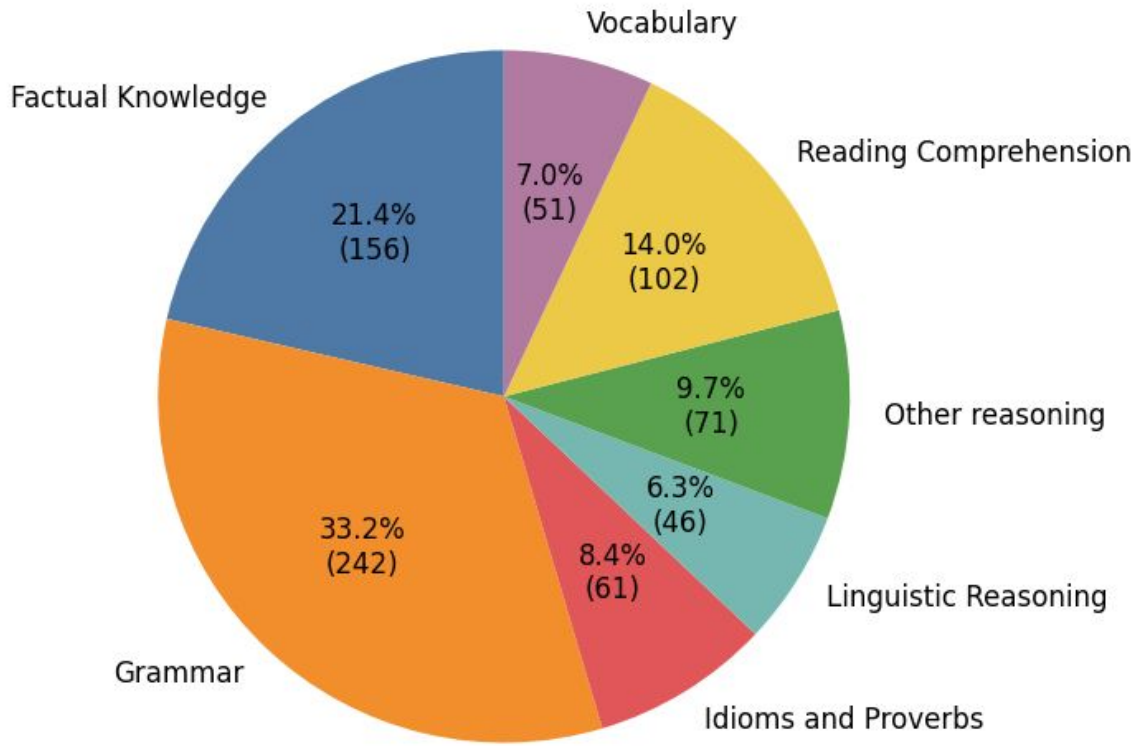

Figure 2: Category-wise distribution of questions in MATA

## 3. Dataset Creation

Motivated by recent efforts (Huang et al., 2023; Tran et al., 2025) that utilize educational materials for benchmark construction, we curate our dataset from Telugu subject textbooks for grades $1$ through $10$ prescribed by state education boards of the two Indian states with Telugu as the official language - Andhra Pradesh and Telangana. Additionally, we also select questions from the Telugu language question papers of various competitive exams that are available online. All content is derived from publicly available government-issued textbooks and exam questions. Note that our approach is slightly different from most of the other work that utilizes educational materials in the sense that we focus on the textbooks/exams of Telugu language rather than content in various subjects (e.g., mathematics, physics, biology etc) written in Telugu.

All questions were manually selected from the textbook lessons or exams. The answers were added by the authors during the annotation process for the school grade material and were available in the key provided for the competitive exam papers. The curation process was carried out by two native Telugu speakers (authors of this paper). Initially, each annotator independently curated a subset of the dataset. Subsequently, both annotators reviewed each other's annotations for accuracy, clarity, linguistic intent, and orthographic correctness (e.g., spelling and punctuation). Any discrepancies were resolved through collaborative discussion to ensure consistency and quality across

[1]pronounced with two syllables: “MAH-tah” and means ”word”

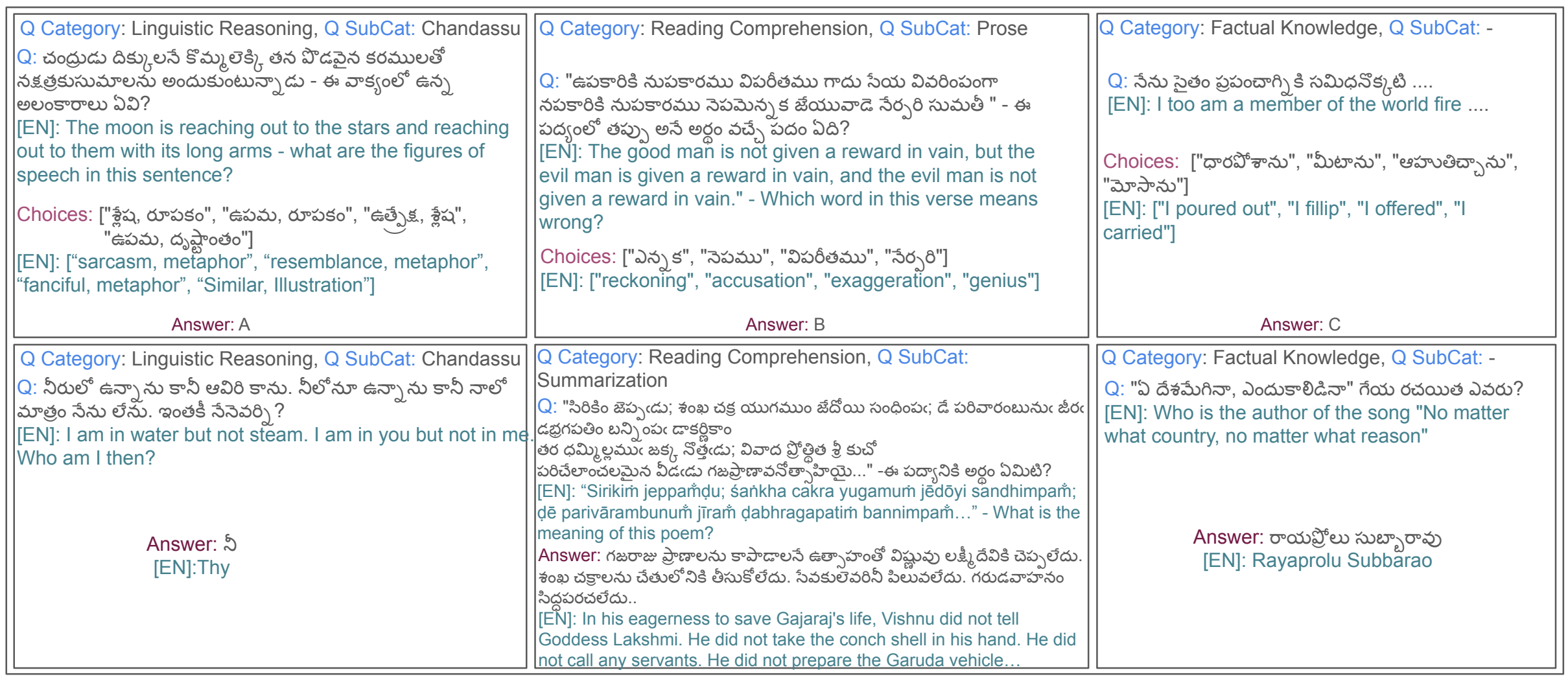

Figure 3: Annotated examples from MATA. Each question includes the original Telugu text, answer choices, and the correct answer. Note: English translations/transliterations are included only for reader accessibility and are not part of the original dataset.

the dataset. The resulting benchmark [2] spans seven high-level categories (see Figure 2) and fifteen subcategories (see Table 3 in Appendix D): grammar, factual knowledge, vocabulary, linguistic reasoning, riddles, and idioms and proverbs Each category includes both multiple-choice and open-ended questions, enabling a broad evaluation of both recognition-based and generative capabilities of LLMs. Approximately $25\%$ of the dataset consists of multiple-choice questions, while the remaining $75\%$ are open-ended. We ensured a range of difficulty by including items from lower grade textbooks (simpler syntax and vocabulary) through high-school texts (advanced grammar and semantics). Question complexity is implicitly stratified through this grade-level sampling. Figure 3 shows some examples from the dataset across categories and question types, with English translation.

## 4. Experimental Setup

We evaluate 11 proprietary and open-weight LLMs on our proposed MATA benchmark using the Inspect evaluation framework [3], with both English (EN) and Telugu (TE) prompts, keeping the default settings for the hyperparameters (*temperature* = $0$).

**Models Evaluated:** Our evaluation includes 11 open-weight and proprietary models. The open-weight models are: `Gemma3-12B`, `Gemma3-27B`, `Qwen3-32B`, `LLama3.3-70B`, and `Sarvam-M-24B`, a LLM trained specifically on Indian languages, while the proprietary models are from `GPT-4o` and `GPT-4.1` from OpenAI, `Sonnet4` from Anthropic, `Grok-3` from XAI, `Gemini2.5-Pro` and `Gemini2.5-Flash` from Google. All models were accessed through API calls to the provider OpenRouter[4] [5]. Most of these models claim multilingual support although the exact list of languages is not explicitly mentioned. Among these models, only `Qwen3` and `Sarvam-M` explicitly list Telugu as one of the supported languages.

**Evaluation Metrics:** As the dataset serves as a diagnostic evaluation resource, we focus on descriptive statistics and qualitative analyses to provide interpretable insights than formal hypothesis testing. We adopt standard evaluation metrics based on the question format. For multiple-choice questions, we report accuracy against the ground truth using Inspect's multiple choice scorer which extracts the answer from the LLM output. For open-ended questions, we incorporate an LLM-as-a-judge approach, where model-generated responses are evaluated against the ground truth answer using `LLaMA3.3-70B` LLM. We also reran the experiments with another judge model (`Gemma3-27B`) as the scorer and present a comparison of the performance in Section 5.3. The judges were prompted to give a binary output (Correct vs Incorrect). To validate the reliability of LLM-based judgments, we also conduct a human evaluation described in Section 5.3 on a subset of data.

Although it is possible the models give partially correct answers sometimes and a binary judgment will penalize such scenarios, it is important to note that partial correctness may not be relevant in all use cases, and would need a deeper qualitative analysis to achieve a better understanding of the model. Hence, we opt for a binary assessment to get a broad picture, and leave a more detailed study of graded assessment as future work for specific application scenarios.

[2] Our dataset is shared publicly under a permissible license

[3] https://inspect.aisi.org.uk/

[4] https://openrouter.ai/

[5] The approximate cost for all the evaluations was 120 USD

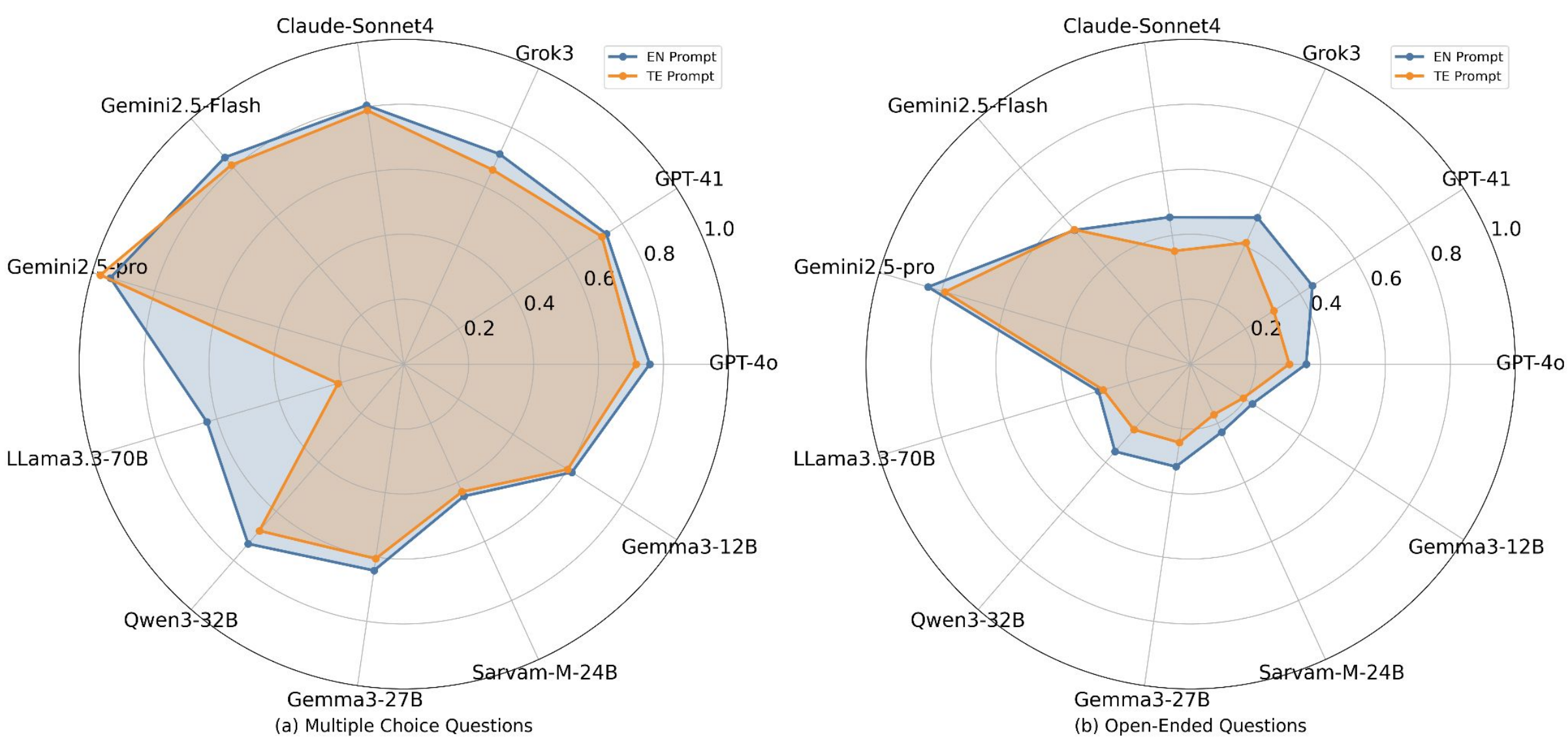

(a) Multiple Choice Questions

(b) Open-Ended Questions

Figure 4: Radar plots comparing model performance (see Tables 4 and 5 in the Appendix for detailed numbers.) on English (EN) and Telugu (TE) prompts across models. Open-ended answers were evaluated using LLama3.3-70B model as the Judge.

## 5. Results

We present a detailed analysis of LLM performance on MATA, followed by additional analysis on the robustness and reliability of the question format and automatic evaluation respectively. All models were evaluated using Telugu (TE) and English (EN) prompts (where the prompts are in either English or Telugu, but the questions are always in Telugu). We present the overall results, organized by question type (multiple-choice and open-ended), followed by a detailed analysis based on model performance across different categories.

### 5.1. LLM Performance on MATA

Figure 4 presents a comparative analysis of model performance on TE and EN prompts for multiple-choice (left) and open-ended (right) questions.

For multiple choice questions, the models performed only slightly better when evaluated with EN prompts compared to TE prompts. Among the closed-source models, `Gemini2.5-Pro` achieved higher accuracy for both TE ($0.974$) and EN ($0.942$) prompts, while `GPT-4.1` showed consistent performance across prompt languages ($0.726$ TE, $0.721$ EN). In contrast, for open-weight models, `Llama3.3-70B` showed a noticeable drop in accuracy with TE prompts ($0.211$ vs $0.632$ EN). A manual analysis revealed that this drastic drop was mainly due to model's failure to consistently follow the expected output format rather than incorrect reasoning. As format compliance is necessary to extract answers automatically from the response, and other LLMs we studied did not encounter this issue, we did not do any model specific adjustments to accomodate the LLama model. Open-ended questions proved to be significantly more challenging, with most models scoring between $0.22$ and $0.45$, substantially lower than multiple-choice performance. This difficulty could stem from two factors: (1) the inherent complexity of the task i.e., generating a free text response rather than selecting from the given options, and (2) the models' tendency to provide incomplete responses in Telugu. For instance, when a question requires listing four items, models often provide only one-to-two items and miss other components, leading to lower evaluation scores despite partial correctness. The gap in performance between multiple-choice and open-ended responses even for the best performing model highlights the limitations of current LLMs in terms of Telugu capabilities.

**Data Contamination:** The relatively high performance of `Gemini2.5-Pro` compared to all other LLMs may lead to the question of data contamination. Note that the answers for the questions from the textbooks were added during the annotation process and are not found in the textbooks themselves. So, while data contamination is possible, it may not be directly from the sources we used for the dataset.

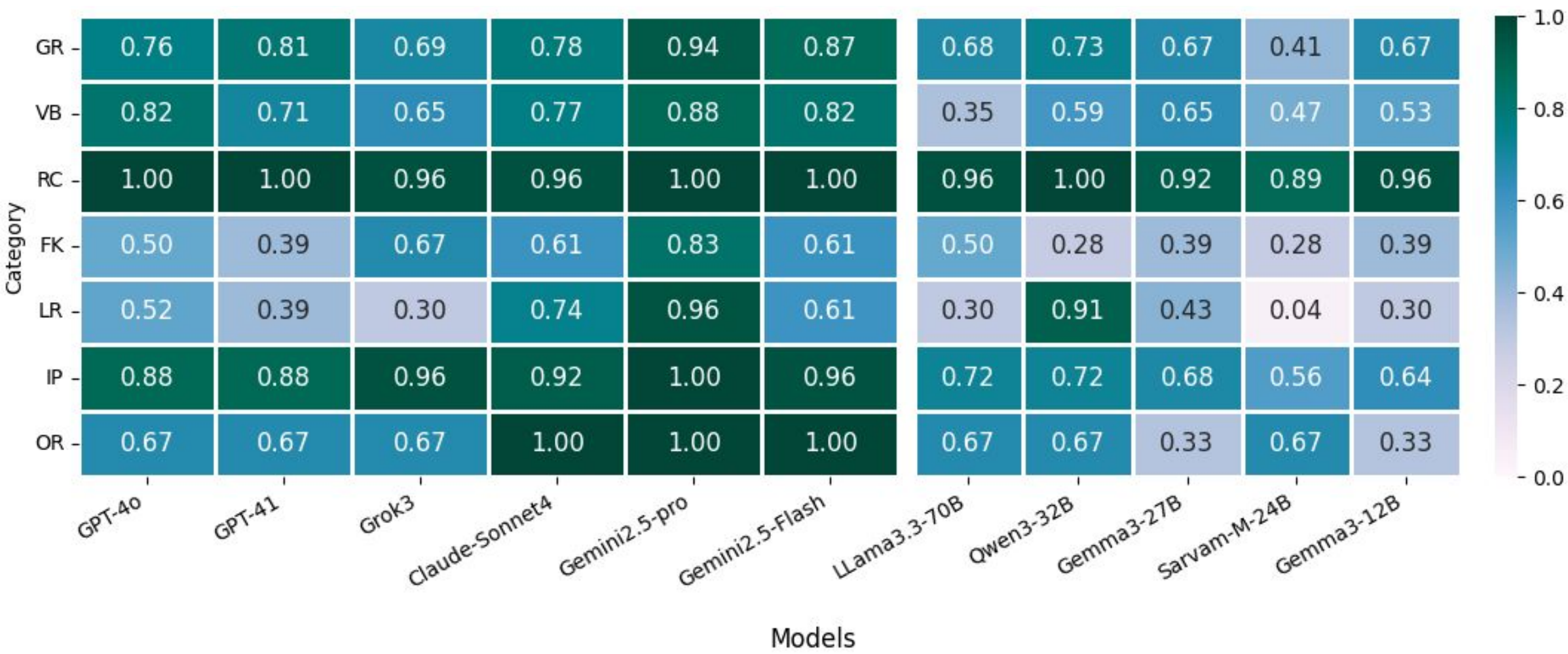

Figure 5: Accuracy of different models on **multiple choice questions** across various language understanding categories using **English** prompt. *G*: Grammar, *V*: Vocabulary, *RC*: Reading Comprehension, *FK*: Factual Knowledge, *IP*: Idioms and Proverbs, *OR*: Other Reasoning. Higher the accuracy, better the performance.

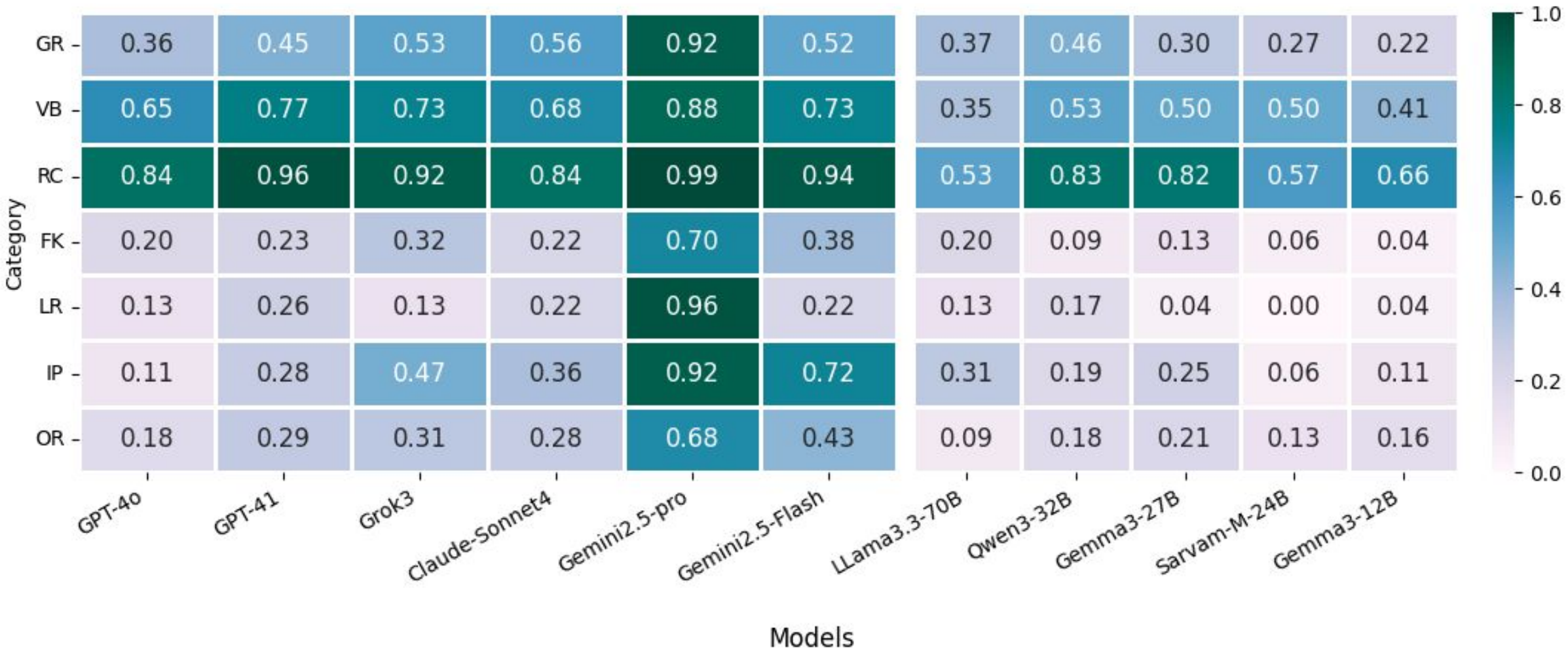

Figure 6: Accuracy of different models on **open-ended questions** across various question categories using **English** prompt. Higher the accuracy, better the performance.

Given the diverse nature of the dataset, we further analyze model performance across the seven question categories - Grammar (GR), Vocabulary (VB), Reading Comprehension (RC), Factual Knowledge (FK), Linguistic Reasoning (LR), Idioms and Proverbs (IP), and Other Reasoning (OR). Considering that all models performed slightly better with English prompts overall, we report all subsequent results with English prompts. (Results for Telugu prompts can be found in the Appendix Tables 5–11). Figures 5 and 6 show the performance of the different LLMs for multiple-choice and open-ended questions respectively across question categories.

For multiple-choice questions, `Gemini2.5-Pro` consistently outperforms other models, achieving over $80\%$ accuracy across all categories. It also reaches $100\%$ accuracy in multiple categories, including RC, IP, and OR. Multiple models also attain $100\%$ accuracy on the RC and OR, suggesting these question types are more tractable for the LLMs. In contrast, the LR category appears particularly challenging, with most models scoring in the $0.3$ - $0.4$ range. These questions require applying phonological rules based on vowel length and syllable structure, which LLMs cannot model explicitly. Exceptions to this trend include `Gemini2.5-Pro` and `Qwen3-32B`, which achieve relatively higher scores. This could be potentially because their training includes phoneme or syllable-aware tokenization. Factual Knowledge is another challenging area, especially for open-weight models. Interestingly, despite `Qwen3-32B`'s strong overall performance, it records its lowest score along with `Sarvam-M-24B` model in this category. This may be because of limited regional data and weak grounding in Telugu cultural entities.

For open-ended questions, while `Gemini2.5-Pro` did not achieve $100\%$ accuracy in any single category, it maintained strong performance overall, with scores exceeding $70\%$ across all categories. Although the model achieved $100\%$ accuracy in the Other Reasoning category for multiple-choice questions, its score dropped to $67\%$ in the open-ended setting, its lowest among all categories. The model's strong MCQ performance likely from pattern matching, where choices act as anchors, but its drop in open-ended questions reveal gaps in figurative understanding and limited generative ability.

Both closed and open-weight models exhibited their weakest performance in the Factual Knowledge category for open-ended questions, suggesting that factual questions, particularly those rooted in Telugu culture, pose a significant challenge (Rohera et al., 2025) in generative response settings. In addition, the Grammar, Linguistic Reasoning, Idioms and Proverbs, and Other Reasoning categories proved difficult for most models, regardless of model type.

Some questions in the Linguistic Reasoning category involve word guesser puzzles, where models must infer a target word based on its presence or absence in a set of word pairs (e.g., *I am a three-letter word; I am comprised of letters that appear in word1 but not in word2, word3 but not word4, what am I?*). These tasks demand fine-grained phonemic and lexical discrimination and the models generally did not perform well. The Idioms and Proverbs category contains expressions rooted in Telugu cultural context and usage, requiring culturally informed language understanding. Similarly, the Other Reasoning category includes riddles and context-dependent logic that rely on regional knowledge and idiomatic conventions for successful interpretation. These categories demand not only linguistic competence but also cultural grounding, and the models consistently struggled to produce accurate responses. These results highlight that while models perform well on certain categories such as reading comprehension (where answer is directly in the provided context) across both question formats, those that require deeper semantic, phonological, or cultural understanding remain particularly challenging. Large scale language specific fine-tuning with specific tasks cover a broad range of tasks grounded in the language and culture could potentially lead to better performance, especially in the reasoning categories.

## 5.2. Robustness Analysis

Between Figures 5 and 6, we notice that there are significant disparities between the performance in multiple-choice versus open-ended questions for the same question category across models. This raises a question: to what extent do these language models truly comprehend the questions? To better understand this, we started with the multiple-choice format, by analyzing the distribution of correct answer positions in the questions. Table 1 shows the distribution. Note that the MATA dataset does not contain *'None of the others'* as a correct answer for any question. As shown in Table 1, there is no strong positional bias within any single category, suggesting that answer distribution is relatively balanced.

| **Category** | **A** | **B** | **C** | **D** |
|---|---|---|---|---|
| Grammar | 23 | 14 | 21 | 20 |
| Vocabulary | 4 | 5 | 6 | 2 |
| Reading Comprehension | 6 | 6 | 7 | 7 |
| Factual Knowledge | 5 | 6 | 4 | 3 |
| Linguistic Reasoning | 7 | 6 | 6 | 4 |
| Idioms and Proverbs | 5 | 4 | 4 | 12 |
| Other Reasoning | 1 | 0 | 1 | 1 |

Table 1: Distribution of multiple-choice answer options (*A*–*D*) across different dataset categories.

Motivated by (Alzahrani et al., 2024; Sánchez-Salido et al., 2025), we designed a set of controlled robustness experiments for multiple-choice questions. In the first setting, we fixed the position of the correct answer such that it always appeared in a specific position (e.g., always the first, second, third, or fourth option) across all questions. In the second setting, the correct answer was replaced with the option 'ఇచ్చిన సమాధానాలలో ఏదీ సరిపోదు' (*'None of the others'*), which was positioned either as choice 'D' (which we call None-at-the-end or NE) or at the original answer's position (which we call None-Random or NR). These modifications were intended to assess whether model performance reflects genuine understanding or reliance on superficial cues and distributional biases. Figure 7 presents the robustness evaluation results for the three categories that had consistently better multiple-choice performance across models: Grammar, Reading Comprehension, and Idioms and Proverbs.

While most models largely retained their accuracy across fixed answer positions (*A*–*D*), certain models exhibited clear preferences or aversions to specific positions. For instance, `Claude-Sonnet 4` achieved $100\%$ accuracy when the correct answer consistently appeared in position *A*, suggesting a positional preference. Similarly, `Qwen3-32B` performed well when the correct answer was in positions *A* or *D* but showed a drop for positions B and C. Comparable trends are observed for `Sarvam-M-24B` and `Gemma3-12B`, where performance varied depending on the position of the correct option. Such positional preferences suggest that some models may rely, at least in part, on shallow heuristics rather than genuine comprehension, raising questions over the generalization capabilities of some LLMs.

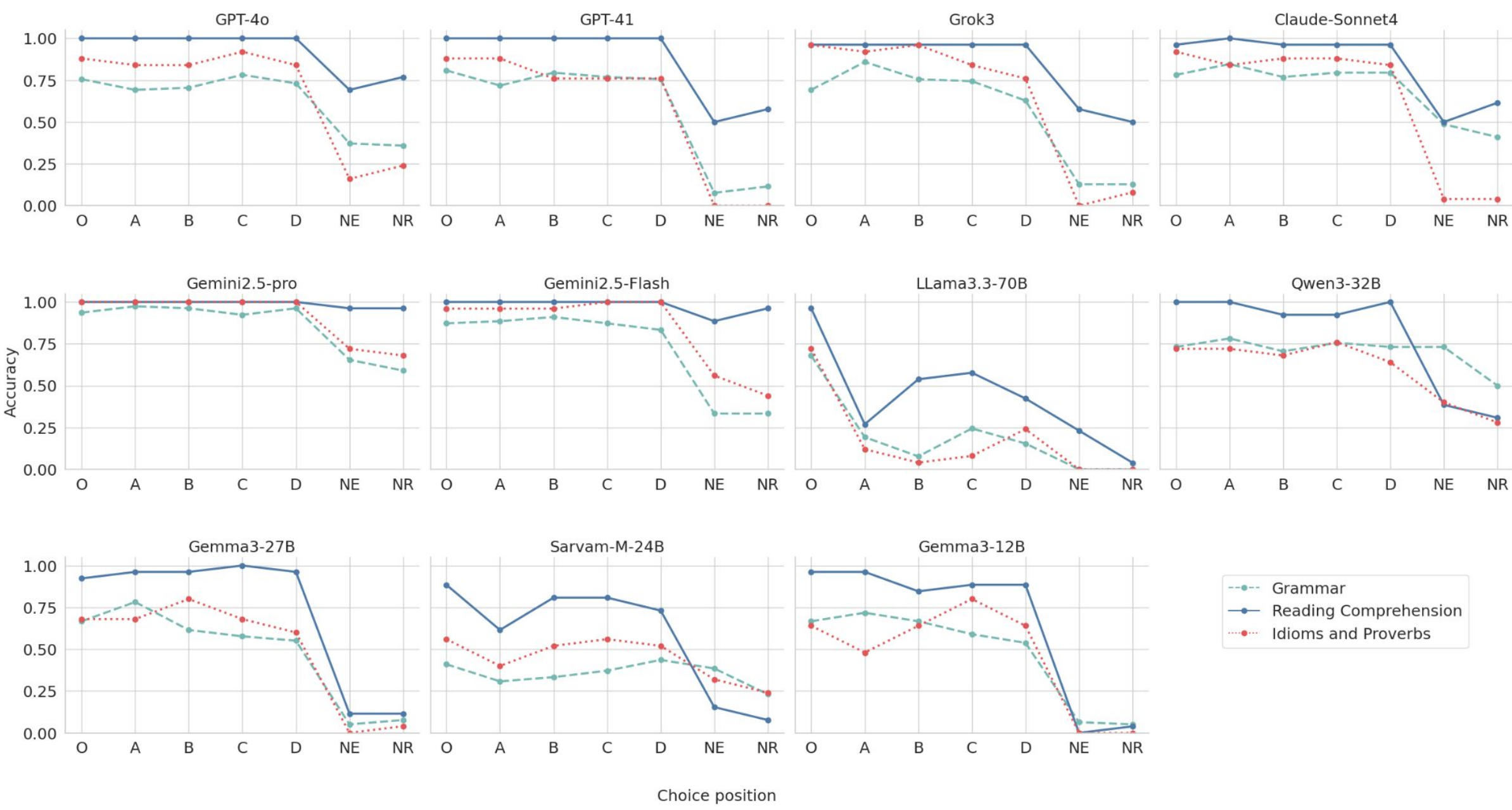

Figure 7: Robustness analysis of multiple-choice question performance under controlled modifications for *Grammar, Reading Comprehension, and Idioms and Proverbs* categories using **EN** prompt: (i) fixing the position of correct answers to A, B, C, or D, and (ii) replacing correct answers with *'None of the others'* placed either at the end (*'NR'*) or at the right answer position (*'NR'*). *O*: Indicates the accuracy received with the choice positions present in the original configuration.

The performance drops observed for `Llama3.3-70B` were again primarily attributable to the model failing to follow instructions. Overall, all models in general exhibited a noticeable decline in accuracy when "None of the others" was included as one of the choices. Interestingly, closed-weight models tended to under perform when this option was placed at the end, whereas open-weight models showed a greater performance drop when the option appeared at a random position. `Gemma3-12B` model 's drop to 0% when "None of the others" was positioned at the end highlights a significant sensitivity to this configuration. This drop in performance when "None of the others" is introduced (which is in-line with Alzahrani et al. (2024) and Sánchez-Salido et al. (2025) showed for English and Spanish) suggests that models struggle to handle negation and out-of-distribution answer patterns, particularly when the structure of the answer set deviates from what they have seen during training or pretraining. Whether this is specifically more severe in low-resource contexts is a question for further investigation in future.

The differential behavior between closed and open-weight models might also need additional analysis. Closed models tend to perform worse when None is placed at the end, possibly because they associate the final position with distractors or low-probability answers in their training distribution. Since many multiple-choice datasets do not use None as a correct answer, these models may implicitly treat such options as implausible, even when correct. Conversely, open-weight models show a more significant drop when None appears at a random position. This could indicate that they rely more heavily on fixed positional expectations or alignment between question types and specific answer choices. The presence of an unexpected or semantically vague option like "None of the others" in a non-final position may introduce ambiguity that disrupts their decision-making process.

## 5.3. Validity of LLM-as-a-Judge Evaluation

All our open-ended question evaluation so far relied on a LLama3.3-70B judge. Does using a different judge model lead us to different conclusions? How reliable is an LLM-judge in a low-resource context like this, compared to human judgements? We explore these questions in this section.

**Variation between the Judge models:** So far, we reported all the open-ended evaluations using LLama3.3-70B judge. However, as mentioned earlier in Section 4, we also repeated all the experiments with Gemma3-27B as an alternate judge. Although there is not much overall difference between the judges across models, there are small to large variations across all models when we break the analysis by question category. (see Tables 7 and 8 in Appendix). Clearly, the choice of the judge can affect what we infer about model performance in such cases. While it is possible to explore other judge models, use LLM-juries instead of single judges, or use fine-tuned judge models for this purpose, further research is in general needed to standardize the use of LLM judges across languages, and we leave it for future exploration.

| LLama3.3-70B Judge | | | | | | | | | |
|---|---|---|---|---|---|---|---|---|---|
| **Category** | **Gemini2.5-pro** | | | **LLama3.3-70B** | | | **Qwen3-32B** | | |
| | JS | A-1 | A-2 | JS | A-1 | A-2 | JS | A-1 | A-2 |
| Vocabulary | 0.882 | 0.933 | 0.933 | 0.353 | 0.833 | 0.917 | 0.529 | 0.889 | 0.889 |
| Linguistic Reasoning | 0.957 | 1.0 | 1.0 | 0.130 | 0.667 | 0.667 | 0.174 | 0.250 | 0.250 |

Table 2: Accuracy of responses from three LLMs as per a LLama judge and two Human Annotators across two question categories, using **English** prompt. *JS*: Judge score. *A1, A2*: Human annotators.

**Comparison with Human Judges:** To understand how LLM judges compare with human judges, we took a randomly selected set of 57 open-ended questions spanning vocabulary and linguistic reasoning categories and the model outputs from three models, and two native speaker human evaluators labeled the outputs as correct/incorrect. We compared these human judgments with the corresponding evaluations by the LLama3.3 judge. Table 2 summarizes the results of this comparison. (See the Table 12 in the Appendix for a similar comparison with a Gemma3 judge).

For two of the three models, both the human annotators reach the same overall accuracy across the categories. There is some disagreement between human annotators only for one case - for the LLama3.3-70B model on the vocabulary questions. This was because of a disagreement only on one data sample of the 34 vocabulary samples. In all cases, though, the judge model's reported accuracies are much lower than human annotator scores, indicating that the judges are potentially penalizing the models more than necessary. Interestingly, the LLama3.3-70B judge model rates generations from itself to be of a much lower quality than those from Gemini-2.5-pro and Qwen3-32B, alleviating concerns of judge models being partial to generations from their own model family. Finally, the model judgments and human judgments overall seem to be closest to each other for the best performing model i.e., Gemini2.5-pro.

Thus, while high scores on the judge model may correlate with high scores from human annotators, low scores from the model judge does not necessarily mean low scores from human judges. We did not specifically analyze for the specific disagreements between model judge and humans or analyze the typical error patterns of the judge model assessments, and we leave that as a question to explore in future.

## 6. Conclusions

We described the creation the MATA evaluation dataset to test the Telugu language capabilities of LLMs, and benchmarked the performance of 11 state-of-the-art LLMs on this dataset. Additionally, we conducted a robustness analysis of the multiple-choice question format and a reliability analysis of the LLM-as-a-judge approach for evaluating open-ended questions. Our main conclusions are summarized below:

1. There are wide performance disparities across categories and question types among all LLMs, although some data subsets such as multiple-choice reading comprehension questions elicit consistently good performance with all LLMs. (Figures 4– 6). Some open-ended reasoning questions are hard even for the largest and presumably most powerful models among the ones we studied.

2. All LLMs are very sensitive to the order and semantics of multiple-choice options and we see large performance drops (over 90% to to 0% in some cases!) after substituting the correct answer with a "none of the others" option (Figure 7).

3. There are variations in the accuracy reported by the two LLM judges, and the judge LLMs tend to assign much lower accuracy to model outputs compared to human annotators. (Section 5.3).

Most of the questions in this dataset sourced from school level textbooks, and hence, can be considered to be a level comprehensible by children and other (human) language learners. Yet, most LLMs struggle to answer them correctly. Asking questions about the language abilities themselves instead of focusing on translated knowledge intensive questions or multiple-choice general knowledge questions in a given language is perhaps essential in understanding the true multilingual capabilities of LLMs. This will require a more mindful development of evaluation datasets and methodology in future. While we focus on one language and a relatively small evaluation set in this paper, we hope that the analysis will provide useful guidelines for those working on evaluating the capabilities of LLMs in other languages.

## Limitations

Since we focused on a manual dataset creation process, the final dataset is not a very large one (729 questions) despite covering a wide range of question categories. We also don't have longer form generative tasks (e.g., summarizing news articles). Further, the dataset does not address specific real-world scenarios. We did not extensively conduct few-shot evaluations or investigate more fine-grained judge assessments instead of a binary correct and incorrect assessment in this paper. Other approaches such as LLMs-as-a-jury or employing fine-tuned judge LLMs instead of generic LLMs for judging were also not explored. Finally, we did not conduct any significance testing, as there are no established guidelines on significance testing for LLM evaluations yet. These limitations are not specific to this particular paper, though, and are of a more general nature, as research focused on LLM benchmarking is still evolving. Despite these limitations, we see this research as a necessary first step towards better evaluation datasets for Telugu and a call for sound methodologies for evaluating LLMs in general, and specifically in low-resource languages, in future.

## Acknowledgments

We thank Krishnapriya Vishnubhotla and Gabriel Bernier-Colborne, and the anonymous reviewers for their helpful feedback. Kranti has been funded by the Bundesministerium für Bildung und Forschung (BMBF, German Federal Ministry of Research), project "COCOBOTS" (01IS21102A) and Deutsche Forschungsgemeinschaft (DFG, German Research Foundation) grant 423217434 ("RECOLAGE").

## References

Muhammad Farid Adilazuarda, Musa Izzanardi Wijanarko, Lucky Susanto, Khumaisa Nur'aini, Derry Wijaya, and Alham Fikri Aji. 2025. Nusaaksara: A multimodal and multilingual benchmark for preserving indonesian indigenous scripts. *CoRR*, abs/2502.18148.

Dhruv Agarwal, Anya Shukla, Sunayana Sitaram, and Aditya Vashistha. 2025. Fluent but culturally distant: Can regional training teach cultural understanding? *CoRR*, abs/2505.21548.

Wafa Alghallabi, Ritesh Thawkar, Sara Ghaboura, Ketan More, Omkar Thawakar, Hisham Cholakkal, Salman Khan, and Rao Muhammad Anwer. 2025. Fann or flop: A multigenre, multiera benchmark for arabic poetry understanding in llms. *arXiv preprint arXiv:2505.18152*.

Emad A. Alghamdi, Reem Masoud, Deema Alnuhait, Afnan Y. Alomairi, Ahmed Ashraf, and Mohamed Zaytoon. 2025. AraTrust: An evaluation of trustworthiness for LLMs in Arabic. In *Proceedings of the 31st International Conference on Computational Linguistics*, pages 8664–8679, Abu Dhabi, UAE. Association for Computational Linguistics.

Norah Alzahrani, Hisham Abdullah Alyahya, Yazeed Alnumay, Sultan Alrashed, Shaykhah Alsubaie, Yousef Almushayqih, Faisal Mirza, Nouf Alotaibi, Nora Al-Twairesh, Areeb Alowisheq, M. Saiful Bari, and Haidar Khan. 2024. When benchmarks are targets: Revealing the sensitivity of large language model leaderboards. In *Proceedings of the 62nd Annual Meeting of the Association for Computational Linguistics (Volume 1: Long Papers), ACL 2024, Bangkok, Thailand, August 11-16, 2024*, pages 13787–13805. Association for Computational Linguistics.

Simone Balloccu, Patrícia Schmidtová, Mateusz Lango, and Ondrej Dusek. 2024. Leak, cheat, repeat: Data contamination and evaluation malpractices in closed-source llms. In *Proceedings of the 18th Conference of the European Chapter of the Association for Computational Linguistics, EACL 2024 - Volume 1: Long Papers, St. Julian's, Malta, March 17-22, 2024*, pages 67–93. Association for Computational Linguistics.

Ayse Aysu Cengiz, Ahmet Kaan Sever, Elif Ecem Ümütlü, Naime Seyma Erdem, Burak Aytan, Büsra Tufan, Abdullah Topraksoy, Esra Darici, and Cagri Toraman. 2025. Evaluating the quality of benchmark datasets for low-resource languages: A case study on turkish. *CoRR*, abs/2504.09714.

Tsz Chung Cheng, Chung Shing Cheng, Chaak Ming Lau, Eugene Tin-Ho Lam, Chun Yat Wong, Hoi On Yu, and Cheuk Hei Chong. 2025. Hkcanto-eval: A benchmark for evaluating cantonese language understanding and cultural comprehension in llms. *CoRR*, abs/2503.12440.

Slawomir Dadas, Malgorzata Grebowiec, Michal Perelkiewicz, and Rafal Poswiata. 2025. Evaluating polish linguistic and cultural competency in large language models. *CoRR*, abs/2503.00995.

Sharvi Endait, Ruturaj Ghatage, Aditya Kulkarni, Rajlaxmi Patil, and Raviraj Joshi. 2025. Indicsquad: A comprehensive multilingual question answering dataset for indic languages. *CoRR*, abs/2505.03688.

Ethnologue. 2025. *What are the top 200 most spoken languages?*

Fan Gao, Cheng Huang, Nyima Tashi, Xiangxiang Wang, Thupten Tsering, Ban Ma-bao, Renzeg Duojie, Gadeng Luosang, Rinchen Dongrub, Dorje Tashi, Xiao Feng, and Yongbin Yu. 2025. TLUE: A tibetan language understanding evaluation benchmark. *CoRR*, abs/2503.12051.

Aryo Pradipta Gema, Joshua Ong Jun Leang, Giwon Hong, Alessio Devoto, Alberto Carlo Maria Mancino, Rohit Saxena, Xuanli He, Yu Zhao, Xiaotang Du, Mohammad Reza Ghasemi Madani, Claire Barale, Robert McHardy, Joshua Harris, Jean Kaddour, Emile Van Krieken, and Pasquale Minervini. 2025. Are we done with MMLU? In *Proceedings of the 2025 Conference of the Nations of the Americas Chapter of the Association for Computational Linguistics: Human Language Technologies (Volume 1: Long Papers)*, pages 5069–5096, Albuquerque, New Mexico. Association for Computational Linguistics.

Dan Hendrycks, Collin Burns, Steven Basart, Andy Zou, Mantas Mazeika, Dawn Song, and Jacob Steinhardt. 2021. Measuring massive multitask language understanding. In *9th International Conference on Learning Representations, ICLR 2021, Virtual Event, Austria, May 3-7, 2021*. OpenReview.net.

Yuzhen Huang, Yuzhuo Bai, Zhihao Zhu, Junlei Zhang, Jinghan Zhang, Tangjun Su, Junteng Liu, Chuancheng Lv, Yikai Zhang, Jiayi Lei, Yao Fu, Maosong Sun, and Junxian He. 2023. C-eval: A multi-level multi-discipline chinese evaluation suite for foundation models. In *Advances in Neural Information Processing Systems 36: Annual Conference on Neural Information Processing Systems 2023, NeurIPS 2023, New Orleans, LA, USA, December 10 - 16, 2023*.

Daeen Kabir, Minhajur Rahman Chowdhury Mahim, Sheikh Shafayat, Adnan Sadik, Arian Ahmed, Eunsu Kim, and Alice Oh. 2025. BLUCK: A benchmark dataset for bengali linguistic understanding and cultural knowledge. *CoRR*, abs/2505.21092.

Chalamalasetti Kranti, Gabriel Bernier-Colborne, Yvan Gauthier, and Sowmya Vajjala. 2025. Test set quality in multilingual LLM evaluation. In *Proceedings of the 5th Workshop on Evaluation and Comparison of NLP Systems*, pages 167–178, Mumbai, India. Association for Computational Linguistics.

Percy Liang, Rishi Bommasani, Tony Lee, Dimitris Tsipras, Dilara Soylu, Michihiro Yasunaga, Yian Zhang, Deepak Narayanan, Yuhuai Wu, Ananya Kumar, Benjamin Newman, Binhang Yuan, Bobby Yan, Ce Zhang, Christian Cosgrove, Christopher D. Manning, Christopher Ré, Diana Acosta-Navas, Drew A. Hudson, Eric Zelikman, Esin Durmus, Faisal Ladhak, Frieda Rong, Hongyu Ren, Huaxiu Yao, Jue Wang, Keshav Santhanam, Laurel J. Orr, Lucia Zheng, Mert Yüksekgönül, Mirac Suzgun, Nathan Kim, Neel Guha, Niladri S. Chatterji, Omar Khattab, Peter Henderson, Qian Huang, Ryan Chi, Sang Michael Xie, Shibani Santurkar, Surya Ganguli, Tatsunori Hashimoto, Thomas Icard, Tianyi Zhang, Vishrav Chaudhary, William Wang, Xuechen Li, Yifan Mai, Yuhui Zhang, and Yuta Koreeda. 2023. Holistic evaluation of language models. *Trans. Mach. Learn. Res.*, 2023.

Chaoqun Liu, Wenxuan Zhang, Jiahao Ying, Mahani Aljunied, Anh Tuan Luu, and Lidong Bing. 2025. Seaexam and seabench: Benchmarking llms with local multilingual questions in southeast asia. In *Findings of the Association for Computational Linguistics: NAACL 2025, Albuquerque, New Mexico, USA, April 29 - May 4, 2025*, pages 6119–6136. Association for Computational Linguistics.

Bernardo Magnini, Roberto Zanoli, Michele Resta, Martin Cimmino, Paolo Albano, Marco Madeddu, and Viviana Patti. 2025. Evalita-llm: Benchmarking large language models on italian. *CoRR*, abs/2502.02289.

Tarek Naous, Michael J. Ryan, Alan Ritter, and Wei Xu. 2024. Having beer after prayer? measuring cultural bias in large language models. In *Proceedings of the 62nd Annual Meeting of the Association for Computational Linguistics (Volume 1: Long Papers), ACL 2024, Bangkok, Thailand, August 11-16, 2024*, pages 16366–16393. Association for Computational Linguistics.

Irene Plaza, Nina Melero, Cristina del Pozo, Javier Conde, Pedro Reviriego, Marina Mayor-Rocher, and María Grandury. 2024. Spanish and LLM benchmarks: is MMLU lost in translation? *CoRR*, abs/2406.17789.

Pritika Rohera, Chaitrali Ginimav, Gayatri Sawant, and Raviraj Joshi. 2025. Better to ask in english? evaluating factual accuracy of multilingual llms in english and low-resource languages. *CoRR*, abs/2504.20022.

Angelika Romanou, Negar Foroutan, Anna Sotnikova, Zeming Chen, Sree Harsha Nelaturu,

Shivalika Singh, Rishabh Maheshwary, Micol Altomare, Mohamed A Haggag, Alfonso Amayuelas, et al. 2024. Include: Evaluating multilingual language understanding with regional knowledge. *arXiv preprint arXiv:2411.19799*.

Angelika Romanou, Negar Foroutan, Anna Sotnikova, Sree Harsha Nelaturu, Shivalika Singh, Rishabh Maheshwary, Micol Altomare, Zeming Chen, Mohamed A Haggag, A Snegha, et al. 2025. Include: Evaluating multilingual language understanding with regional knowledge. In *The Thirteenth International Conference on Learning Representations*.

Eva Sánchez-Salido, Julio Gonzalo, and Guillermo Marco. 2025. None of the others: a general technique to distinguish reasoning from memorization in multiple-choice LLM evaluation benchmarks. *CoRR*, abs/2502.12896.

Hyopil Shin, Sangah Lee, Dongjun Jang, Wooseok Song, Jaeyoon Kim, Chaeyoung Oh, Hyemi Jo, Youngchae Ahn, Sihyun Oh, Hyohyeong Chang, Sunkyoung Kim, and Jinsik Lee. 2025. Kobalt: Korean benchmark for advanced linguistic tasks. *CoRR*, abs/2505.16125.

Serry Sibaee, Omer Nacar, Adel Ammar, Yasser Al-Habashi, Abdulrahman Al-Batati, and Wadii Boulila. 2025. From guidelines to practice: A new paradigm for arabic language model evaluation. *arXiv preprint arXiv:2506.01920*.

Harman Singh, Nitish Gupta, Shikhar Bharadwaj, Dinesh Tewari, and Partha Talukdar. 2024a. IndicGenBench: A multilingual benchmark to evaluate generation capabilities of LLMs on Indic languages. In *Proceedings of the 62nd Annual Meeting of the Association for Computational Linguistics (Volume 1: Long Papers)*, pages 11047–11073, Bangkok, Thailand. Association for Computational Linguistics.

Shivalika Singh, Angelika Romanou, Clémentine Fourrier, David I. Adelani, Jian Gang Ngui, Daniel Vila-Suero, Peerat Limkonchotiwat, Kelly Marchisio, Wei Qi Leong, Yosephine Susanto, Raymond Ng, Shayne Longpre, Wei-Yin Ko, Madeline Smith, Antoine Bosselut, Alice Oh, Andre F. T. Martins, Leshem Choshen, Daphne Ippolito, Enzo Ferrante, Marzieh Fadaee, Beyza Ermis, and Sara Hooker. 2024b. *Global MMLU: Understanding and Addressing Cultural and Linguistic Biases in Multilingual Evaluation*.

Yosephine Susanto, Adithya Venkatadri Hulagadri, Jann Montalan, Jian Gang Ngui, Xian Bin Yong, Wei Qi Leong, Hamsawardhini Rengarajan, Peerat Limkonchotiwat, Yifan Mai, and William-Chandra Tjhi. 2025. SEA-HELM: southeast asian holistic evaluation of language models. *CoRR*, abs/2502.14301.

Khanh-Tung Tran, Barry O'Sullivan, and Hoang D. Nguyen. 2025. Irlbench: A multi-modal, culturally grounded, parallel irish-english benchmark for open-ended LLM reasoning evaluation. *CoRR*, abs/2505.13498.

Sshubam Verma, Mohammed Safi Ur Rahman Khan, Vishwajeet Kumar, Rudra Murthy, and Jaydeep Sen. 2024. Milu: A multi-task indic language understanding benchmark. *arXiv preprint arXiv:2411.02538*.

Shangqing Zhao, Yuhao Zhou, Yupei Ren, Zhe Chen, Chenghao Jia, Fang Zhe, Zhaogaung Long, Shu Liu, and Man Lan. 2025. Fùxì: A benchmark for evaluating language models on ancient chinese text understanding and generation. *CoRR*, abs/2503.15837.

## A. Dataset Sources

As mentioned in Section 3, we curated our dataset from multiple educational sources. Specifically, the data was collected from publicly available class textbooks prescribed by the Andhra Pradesh and Telangana State Boards in India, as well as from a range of competitive examination question papers sourced from online platforms. All sources used are publicly available and intended for educational use. The sources for our dataset are listed below.

1. Andhra Pradesh State Board textbooks (Grades 1–10): `https://www.apteachers.in/`
2. Telangana State Board textbooks (Grades 1–10): `https://scert.telangana.gov.in/Displaycontent.aspx?encry=ammkNW4/gx+NeApstGPX+A==`
3. Competitive Exam Questions: `https://education.sakshi.com/en/group-1/previous-papers`
4. News Paper Resources: `https://www.eenadu.net/telugu-article/kids-st`

## B. Dataset Examples

Figure 3 presents representative examples (along with an English translation) from our curated dataset. Each data instance includes the following fields: category (e.g., Linguistic Reasoning, Reading Comprehension, Factual Knowledge etc.), sub-category (e.g., Chandassu, Riddles, Summarization etc.), the question text in Telugu, a list of choices (for multiple-choice questions), and the correct choice (A/B/C/D). In addition to multiple-choice questions, our dataset also contains open-ended questions, which require free-form textual responses instead of selecting from predefined options.

| Category | Sub-Category | Question Type | |
|---|---|---|---|
| | | Multiple-Choice | Open-Ended |
| Grammar | Word Formation | 50 | 83 |
| | Sentence Formation | 14 | 24 |
| | Compounding | 9 | 51 |
| | Other (punctuation, PoS) | 5 | 6 |
| Vocabulary | - | 17 | 34 |
| Reading Comprehension | Prose | 2 | 21 |
| | Summarization (Poetry) | 24 | 55 |
| Factual Knowledge | - | 18 | 138 |
| Linguistic Reasoning | Word Guesser | - | 17 |
| | Chandassu | 23 | 6 |
| Idioms and Proverbs | - | 25 | 36 |
| Other Reasoning | Riddles | 3 | 68 |
| Total | - | 190 | 539 |

Table 3: Dataset Stats

## C. Question Types

Our dataset consists of two types of questions: multiple choice and open-ended. Multiple Choice Questions (MCQs) provide a set of four predefined answer options and only one correct choice. These are commonly used in educational settings and allow for automatic evaluation. In contrast, Open-Ended Questions (OEQs) require models to produce free-form textual responses. These questions often demand descriptive or explanatory answers and are inherently more challenging to evaluate automatically. Both MCQs and OEQs are annotated with a grade-level difficulty ranging from Grade 1 to Grade 10. The inclusion of both formats ensures a diverse assessment of language understanding, reasoning, and generative abilities in Telugu.

## D. Dataset Structure

Our dataset is organized into multiple categories (see Table 3) that reflect distinct linguistic capabilities. Each question is annotated with one of the following high-level categories:

The **Grammar** category evaluates how well LLMs can understand and apply Telugu grammatical [6] rules. Questions in this category involve compound words formed through morphological rules (sandhi) and semantic combination of words (samasam), as well as tasks involving sentence formation and word construction by correctly ordering words or letters. This category also includes items on punctuation usage and part-of-speech (PoS) identification.

The **Factual Knowledge** category assesses the model's familiarity with Telugu culture, literature, notable authors, and classical texts. These questions are designed to evaluate the extent to which LLMs are grounded in culturally specific knowledge relevant to Telugu.

The **Vocabulary** category focuses on evaluating the model's lexical understanding. It includes questions on word meanings, contextual interpretations, synonyms, and semantic similarities, thereby testing the model's knowledge of both surface-level and nuanced vocabulary.

The **Linguistic Reasoning** category covers more advanced linguistic constructs, with a focus on chandassu [7], a prosodic feature similar to metre in English poetry. This category tests the model's ability to reason about and follow phonological and structural constraints inherent in classical Telugu poetic forms.

The **Reading Comprehension** category includes questions based on prose and poetry excerpts, especially classical Telugu poems (padyalu). Tasks involve answering interpretive questions and summarizing the given passages, thus testing both comprehension and generative abilities.

The **Idioms and Proverbs** category includes idiomatic expressions and culturally rooted proverbs that are common in Telugu. This category evaluates the model's ability to interpret figurative language and understand expressions that carry meanings beyond their literal composition.

Finally, the **Other Reasoning** category consists of riddles and evaluates the model's ability to interpret metaphorical and symbolic language.

[6] https://en.wikipedia.org/wiki/Telugu_grammar

[7] https://en.wikipedia.org/wiki/Telugu_grammar#Chandassu_or_Telugu_prosody

TEMPLATE D.0.1

Answer the following multiple choice question. The entire content of your response should be of the following format: 'ANSWER: $LETTER' (without quotes) where LETTER is one of A,B,C,D.

Figure 8: Default English Prompt template for the multiple choice questions from the Inspect Framework.

TEMPLATE D.0.2
మీరు నిష్ణాతులైన తెలుగు పండితులు మరియు ప్రశ్నలను మూల్యాంకనం చేయడంలో అనుభవజ్ఞులు.

క్రింద ఒక ప్రశ్న, సరియైన సమాధానం, మూల్యాంకనం చేయవలసిన సమాధానం ఉన్నాయి.

మీరు మూల్యాంకనం చేయవలసిన సమాధానం, సరియైన సమాధానానికి ఎంత దగ్గరగా ఉందో పోల్చి చెప్పాలి.
రెండు సమాధానాలు ఒకటే లేదా బాగా దగ్గరగా ఉంటే: "అవును" అని చెప్పండి. అలా కాని పక్షంలో "లేదు" అని చెప్పండి.

నియమాలు:
1. ఏ అదనపు వివరణలు లేదా వ్యాఖ్యలు అందించవద్దు. పైన ఇచ్చిన పద్ధతిలో మాత్రమే స్పందించండి.

ప్రశ్న:
$QUESTION
సరియైన సమాధానం:
$GROUND_TRUTH
మూల్యాంకనం చేయవలసిన సమాధానం:
$MODEL_RESPONSE

Figure 9: Telugu Prompt template for using LLM as a judge in the evaluation.

These items typically present a set of descriptive or metaphorical clues, from which the model must infer the correct answer. This category tests a model's reasoning under ambiguity, cultural alignment, and ability to decode non-literal language.

# E. Prompt Templates

We use two styles of prompt templates for our experiments: English (EN) Prompt and Telugu (TE) Prompt. The EN Prompt follows the default template provided by the Inspect framework (see Figure 8). The TE Prompt (see Figure 10 ) follows standard zero-shot prompting structure comprising a system message, task description, and details about the expected response format. We follow the same distinction to Judge Prompts, which are used for prompting models to evaluate the model generated responses. The EN Judge Prompt uses the default format from the Inspect framework (see Figure 11). For Telugu, we designed a custom TE Judge Prompt (see Figure 9) that mirrors the structure of the TE task prompt, ensuring that the evaluation criteria and response expectations are explained in Telugu.

TEMPLATE E.0.1
మీ పని క్రింద ఇచ్చిన ప్రశ్నకు సరైన జవాబును ఎంపిక చేయడం.

సూచనలు:
1. మీరు నాలుగు ఎంపికల నుండి ఒక సమాధానాన్ని మాత్రమే ఎంచుకోవాలి మరియు సరైన సమాధానానికి సంబంధించిన అక్షరంతో (A, B, C, లేదా D) స్పందించాలి.

2. ఏ అదనపు వివరణలు లేదా వ్యాఖ్యలు అందించవద్దు. సరైన సమాధానం యొక్క అక్షరాన్ని మాత్రమే ఇవ్వండి.

3. మీరు సమాధానం ఖచ్చితంగా తెలియకపోతే, ప్రశ్నపై మీ అవగాహన ఆధారంగా అత్యంత సముచితంగా అనిపించే ఎంపికను ఎంచుకోండి.

నియమాలు:
1. పైన ఇచ్చిన పద్ధతిలో మాత్రమే స్పందించండి.

2. జవాబు తప్పనిసరిగా ఇచ్చిన నాలుగు సమాధానాలలోది మాత్రమే అయి ఉండాలి.

జవాబు 'ANSWER: A, B, C, D' ఈ నాలుగు ఆప్షన్స్ లలో ఒకటి అయ్యి ఉండాలి.

Figure 10: Telugu Prompt template for the multiple choice questions.

# F. Results

## F.1. Performance Based On Question Type

Table 4 and Table 5 present model performance by question type (Multiple Choice and Open-Ended) and prompt style (EN and TE). For open-ended questions, we report evaluation scores using both `LLaMA3-70B` and `Gemma3-27B` as judges.

With the English prompt, `Gemini2.5-pro` achieved the higher score for multiple choice questions ($0.942$), and also performed strongly on open-ended questions, with scores of $0.843$ (LLaMA Judge) and $0.819$ (Gemma Judge). Among open-weight models, `Qwen3-32B` showed the better performance with the English prompt, with scores of $0.732$ (MCQ), $0.357$ (LLaMA Judge), and $0.337$ (Gemma Judge).

When evaluated with the Telugu prompt, `Gemini2.5-pro` achieved slightly higher scores:

TEMPLATE E.0.2

**TASK**

You are comparing a submitted answer to an expert answer on a given question. Here is the data:

BEGIN DATA

************

[Question]:

$QUESTION

***********

Expert

$GROUNDTRUTH_ANSWER

***********

Submission:

$GENMODEL_RESPONSE

***********

END DATA

**INSTRUCTIONS:**

Compare the factual content of the submitted answer with the expert answer. Ignore any differences in style, grammar, or punctuation.

Does the submission contain the content in the expert answer?

After assessing the submitted answer, reply with 'GRADE: $LETTER' (without quotes) where LETTER is one of CI. Please choose ONE option for the grade: either C for correct answers, or I for incorrect answers.

For example, after reviewing a correct answer you might write 'GRADE: C' or after reviewing an incorrect answer you might write 'GRADE: I'

First, write out in a step by step manner your reasoning about the criterion to be sure that your conclusion is correct. Avoid simply stating the correct answers at the outset. Then, end with your answer formatted as 'GRADE: $LETTER' (without quotes) where LETTER is one of CI.

Figure 11: Default English Prompt template for using LLM as a judge in the evaluation in Inspect Framework.

$0.974$ (MCQ), $0.824$ (LLaMA Judge), and $0.789$ (Gemma Judge). In the case of open-weight models, `Qwen3-32B` again performed better on multiple choice questions ($0.679$), while `LLaMA3.3-70B` scored higher on open-ended responses with $0.293$ (LLaMA Judge) and $0.281$ (Gemma Judge).

The performance gap observed between EN and TE prompts, particularly for open-weight models, suggests an interesting direction for future investigation.

## F.2. Performance Based On Category

We further analyzed model performance on both MCQ and OEQ tasks based on the linguistic category of each question, across prompt styles. Table 6 reports category-wise MCQ accuracy with EN prompt. Several models: `GPT-4o`, `GPT-41`, `Gemini2.5-Flash`, `Gemini2.5-pro`, and `Qwen3-32B`, achieved 100% accuracy in the Reading Comprehension category. Similarly, in the Other Reasoning category, models like Claude Sonnet, Gemini2.5-Flash, and Gemini2.5-pro also reached $100\%$ scores. Overall, higher accuracy scores were observed in categories such as Reading Comprehension, Idioms and Proverbs, Grammar, and Other Reasoning.

Table 9 presents the corresponding category-wise MCQ results with the TE prompt. Again, models such as GPT-4o, GPT-41, Gemini2.5-Flash, and Gemini2.5-pro achieved $100\%$ accuracy in Reading Comprehension category. High scores were also observed for Idioms and Proverbs and Other Reasoning, for the Gemini models. Broadly, the categories where models performed well with the EN prompt also saw similar results with the TE prompt, indicating consistent behavior across prompt styles for certain categories of the data.

We extended the category-wise analysis to open-ended questions, using both LLaMA3.3-70B and Gemma3-27B as judges with the EN prompt. Tables 7 and 8 present these results. Categories such as Factual Knowledge, Linguistic Reasoning, Idioms and Proverbs, and Other Reasoning consistently yielded lower scores across most models. Nevertheless, Gemini2.5-Pro achieved relatively better performance for these challenging categories, with scores ranging from $0.68$ to $0.96$. Under the Gemma3-27B judge, scores in these categories remained similarly low; Also in the Linguistic Reasoning category, models such as GPT-4o, Gemma3-27B, and Gemma3-12B received scores of 0, underscoring the complexity of this category.

Tables 10 and 11 show the corresponding results for TE prompts. Under this setting, Gemini2.5-pro achieved a score of $1.0$ in the Linguistic Reasoning category. However, for most other models, the performance trends remained consistent with the EN prompt setting, and reports lower scores in Factual Knowledge and Idioms and Proverbs—highlighting persistent challenges across prompt styles.

| Model | Question Type | | |
|---|---|---|---|
| | Multiple-Choice | Open-Ended | |
| | | Gemma3-27B as a Judge | Llama3.3-70B as a Judge |
| GPT-4o | 0.758(0.031) | 0.337(0.020) | 0.356(0.021) |
| GPT-41 | 0.721(0.033) | 0.431(0.021) | 0.446(0.021) |
| Grok3 | 0.711(0.033) | 0.448(0.021) | 0.496(0.022) |
| Claude-Sonnet4 | 0.805(0.029) | 0.465(0.021) | 0.457(0.022) |
| Gemini2.5-Flash | 0.842(0.027) | 0.550(0.021) | 0.546(0.021) |
| Gemini2.5-pro | **0.942**(0.017) | **0.819**(0.017) | **0.843**(0.016) |
| LLama3.3-70B | 0.632(0.035) | 0.278(0.019) | 0.296(0.020) |
| Qwen3-32B | 0.732(0.032) | 0.337(0.020) | 0.357(0.021) |
| Gemma3-27B | 0.642(0.035) | 0.319(0.020) | 0.319(0.020) |
| Sarvam-M-24B | 0.447(0.036) | 0.226(0.018) | 0.230(0.018) |
| Gemma3-12B | 0.616(0.035) | 0.215(0.018) | 0.226(0.018) |

Table 4: Performance of different models on multiple choice and open-ended questions across various models using **English** prompt. Accuracy and standard deviation in parantheses. Higher the accuracy the better the performance.

| Model | Question Type | | |
|---|---|---|---|
| | Multiple-Choice | Open-Ended | |
| | | Gemma3-27B as a Judge | Llama3.3-70B as a Judge |
| GPT-4o | 0.716(0.033) | 0.309(0.020) | 0.304(0.020) |
| GPT-41 | 0.726(0.032) | 0.333(0.020) | 0.304(0.020) |
| Grok3 | 0.658(0.035) | 0.407(0.021) | 0.411(0.021) |
| Claude-Sonnet4 | 0.789(0.030) | 0.376(0.021) | 0.352(0.021) |
| Gemini2.5-Flash | 0.811(0.029) | 0.570(0.021) | 0.548(0.021) |
| Gemini2.5-pro | **0.974**(0.017) | **0.824**(0.016) | **0.789**(0.018) |
| LLama3.3-70B | 0.211(0.030) | 0.293(0.020) | 0.281(0.019) |
| Qwen3-32B | 0.679(0.034) | 0.269(0.020) | 0.267(0.020) |
| Gemma3-27B | 0.605(0.036) | 0.250(0.019) | 0.244(0.019) |
| Sarvam-M-24B | 0.432(0.036) | 0.167(0.016) | 0.200(0.017) |
| Gemma3-12B | 0.600(0.036) | 0.189(0.017) | 0.193(0.017) |

Table 5: Performance of different models on multiple choice and open-ended questions across various models using **Telugu** prompt. Accuracy and standard deviation in parantheses. Higher the accuracy the better the performance.

| Model | G | V | RC | FK | LR | IP | OR |
|---|---|---|---|---|---|---|---|
| GPT-4o | 0.756 | 0.824 | **1.0** | 0.500 | 0.522 | 0.880 | 0.667 |
| GPT-41 | 0.808 | 0.706 | **1.0** | 0.389 | 0.391 | 0.880 | 0.667 |
| Grok3 | 0.692 | 0.647 | 0.962 | 0.667 | 0.304 | 0.960 | 0.667 |
| Claude-Sonnet4 | 0.782 | 0.765 | 0.962 | 0.611 | 0.611 | 0.920 | **1.0** |
| Gemini2.5-Flash | 0.872 | 0.824 | **1.0** | 0.611 | 0.609 | 0.960 | **1.0** |
| Gemini2.5-pro | **0.936** | **0.882** | **1.0** | **0.833** | **0.957** | **1.0** | **1.0** |
| LLama3.3-70B | 0.679 | 0.353 | 0.962 | 0.500 | 0.304 | 0.720 | 0.667 |
| Qwen3-32B | 0.731 | 0.588 | **1.0** | 0.278 | 0.913 | 0.720 | 0.667 |
| Gemma3-27B | 0.667 | 0.647 | 0.923 | 0.389 | 0.435 | 0.680 | 0.333 |
| Sarvam-M-24B | 0.410 | 0.471 | 0.885 | 0.278 | 0.043 | 0.560 | 0.667 |
| Gemma3-12B | 0.667 | 0.529 | 0.962 | 0.389 | 0.304 | 0.640 | 0.333 |

Table 6: Accuracy of different models on multiple choice questions across various language understanding categories using **English** prompt. *G*: Grammar, *V*: Vocabulary, *RC*: Reading Comprehension, *FK*: Factual Knowledge, *IP*: Idioms and Proverbs, *OR*: Other Reasoning.

| **Model** | **G** | **V** | **RC** | **FK** | **LR** | **IP** | **OR** |
|---|---|---|---|---|---|---|---|
| GPT-4o | 0.360 | 0.647 | 0.844 | 0.196 | 0.130 | 0.111 | 0.176 |
| GPT-41 | 0.451 | 0.765 | 0.961 | 0.225 | 0.261 | 0.278 | 0.294 |
| Grok3 | 0.53 | 0.735 | 0.922 | 0.319 | 0.130 | 0.472 | 0.309 |
| Claude-Sonnet4 | 0.561 | 0.676 | 0.844 | 0.217 | 0.217 | 0.361 | 0.279 |
| Gemini2.5-Flash | 0.518 | 0.735 | 0.935 | 0.384 | 0.217 | 0.722 | 0.426 |
| Gemini2.5-pro | **0.921** | **0.882** | **0.987** | **0.703** | **0.957** | **0.917** | **0.676** |
| LLama3.3-70B | 0.366 | 0.353 | 0.532 | 0.196 | 0.130 | 0.306 | 0.088 |
| Qwen3-32B | 0.457 | 0.529 | 0.831 | 0.087 | 0.174 | 0.194 | 0.176 |
| Gemma3-27B | 0.305 | 0.500 | 0.818 | 0.130 | 0.043 | 0.250 | 0.206 |
| Sarvam-M-24B | 0.268 | 0.500 | 0.571 | 0.058 | 0.0 | 0.056 | 0.132 |
| Gemma3-12B | 0.220 | 0.412 | 0.662 | 0.036 | 0.043 | 0.111 | 0.162 |

Table 7: Accuracy of different models on open-ended questions using LLama3.3-70B as judge across various language understanding categories using **English** prompt. *G*: Grammar, *V*: Vocabulary, *RC*: Reading Comprehension, *FK*: Factual Knowledge, *IP*: Idioms and Proverbs, *OR*: Other Reasoning.

| **Model** | **G** | **V** | **RC** | **FK** | **LR** | **IP** | **OR** |
|---|---|---|---|---|---|---|---|
| GPT-4o | 0.372 | 0.588 | 0.753 | 0.167 | 0.0 | 0.111 | 0.235 |
| GPT-41 | 0.463 | 0.794 | 0.870 | 0.239 | 0.217 | 0.250 | 0.235 |
| Grok3 | 0.482 | 0.824 | 0.831 | 0.326 | 0.087 | 0.278 | 0.206 |
| Claude-Sonnet4 | 0.549 | 0.794 | 0.857 | 0.225 | 0.174 | 0.472 | 0.235 |
| Gemini2.5-Flash | 0.561 | **0.882** | 0.948 | 0.384 | 0.087 | 0.611 | 0.368 |
| Gemini2.5-pro | **0.890** | **0.882** | **0.974** | **0.710** | **0.826** | **0.778** | **0.676** |
| LLama3.3-70B | 0.329 | 0.412 | 0.558 | 0.167 | 0.087 | 0.250 | 0.074 |
| Qwen3-32B | 0.494 | 0.647 | 0.649 | 0.080 | 0.261 | 0.167 | 0.088 |
| Gemma3-27B | 0.274 | 0.588 | 0.779 | 0.116 | 0.0 | 0.417 | 0.235 |
| Sarvam-M-24B | 0.256 | 0.500 | 0.481 | 0.072 | 0.043 | 0.167 | 0.132 |
| Gemma3-12B | 0.207 | 0.500 | 0.623 | 0.065 | 0.0 | 0.056 | 0.088 |

Table 8: Accuracy of different models on open-ended questions using Gemma3-27B as judge across various language understanding categories using **English** prompt. *G*: Grammar, *V*: Vocabulary, *RC*: Reading Comprehension, *FK*: Factual Knowledge, *IP*: Idioms and Proverbs, *OR*: Other Reasoning.

| **Model** | **G** | **V** | **RC** | **FK** | **LR** | **IP** | **OR** |
|---|---|---|---|---|---|---|---|
| GPT-4o | 0.705 | 0.647 | **1.0** | 0.444 | 0.478 | 0.920 | 0.667 |
| GPT-41 | 0.782 | 0.824 | **1.0** | 0.444 | 0.261 | 0.840 | 0.667 |
| Grok3 | 0.692 | 0.588 | 0.923 | 0.444 | 0.174 | 0.920 | 0.667 |
| Claude-Sonnet4 | 0.795 | 0.765 | **1.0** | 0.444 | 0.696 | 0.920 | 0.667 |
| Gemini2.5-Flash | 0.833 | 0.882 | **1.0** | 0.667 | 0.348 | **1.0** | **1.0** |
| Gemini2.5-pro | **0.962** | **0.941** | **1.0** | **0.944** | **1.0** | **1.0** | **1.0** |
| LLama3.3-70B | 0.192 | 0.353 | 0.500 | 0.111 | 0.0 | 0.160 | 0.0 |
| Qwen3-32B | 0.756 | 0.294 | 0.923 | 0.278 | 0.739 | 0.680 | 0.667 |
| Gemma3-27B | 0.667 | 0.647 | 0.808 | 0.444 | 0.174 | 0.720 | 0.333 |
| Sarvam-M-24B | 0.462 | 0.353 | 0.692 | 0.389 | 0.043 | 0.480 | 0.667 |
| Gemma3-12B | 0.654 | 0.471 | 0.923 | 0.278 | 0.348 | 0.680 | 0.333 |

Table 9: Accuracy of different models on multiple choice questions across various language understanding categories using **Telugu** prompt. *G*: Grammar, *V*: Vocabulary, *RC*: Reading Comprehension, *FK*: Factual Knowledge, *IP*: Idioms and Proverbs, *OR*: Other Reasoning.

| Model | G | V | RC | FK | LR | IP | OR |
|---|---|---|---|---|---|---|---|
| GPT-4o | 0.299 | 0.559 | 0.740 | 0.181 | 0.043 | 0.083 | 0.147 |
| GPT-41 | 0.299 | 0.618 | 0.753 | 0.138 | 0.130 | 0.056 | 0.176 |
| Grok3 | 0.402 | 0.676 | 0.779 | 0.275 | 0.130 | 0.333 | 0.294 |
| Claude-Sonnet4 | 0.396 | 0.647 | 0.727 | 0.188 | 0.043 | 0.056 | 0.265 |
| Gemini2.5-Flash | 0.543 | **0.794** | 0.922 | 0.413 | 0.13 | 0.583 | 0.412 |
| Gemini2.5-pro | **0.829** | **0.794** | **0.974** | **0.659** | **1.0** | **0.750** | **0.691** |
| LLama3.3-70B | 0.305 | 0.382 | 0.597 | 0.188 | 0.087 | 0.333 | 0.044 |
| Qwen3-32B | 0.378 | 0.529 | 0.545 | 0.036 | 0.304 | 0.028 | 0.132 |
| Gemma3-27B | 0.274 | 0.441 | 0.597 | 0.094 | 0.043 | 0.083 | 0.147 |
| Sarvam-M-24B | 0.268 | - | 0.424 | 0.043 | 0.167 | 0.083 | - |
| Gemma3-12B | 0.213 | 0.382 | 0.468 | 0.058 | 0.043 | 0.083 | 0.118 |

Table 10: Accuracy of different models on open-ended questions using LLama3.3-70B as judge across various language understanding categories using **Telugu** prompt. *G*: Grammar, *V*: Vocabulary, *RC*: Reading Comprehension, *FK*: Factual Knowledge, *IP*: Idioms and Proverbs, *OR*: Other Reasoning. For the Sarvam-M model, responses were not received from OpenRouter for two categories despite multiple runs; hence, results for those categories are not reported.

| Model | G | V | RC | FK | LR | IP | OR |
|---|---|---|---|---|---|---|---|
| GPT-4o | 0.335 | 0.618 | 0.701 | 0.159 | 0.087 | 0.194 | 0.088 |
| GPT-41 | 0.341 | 0.618 | 0.766 | 0.181 | 0.0 | 0.167 | 0.191 |
| Grok3 | 0.402 | 0.765 | 0.792 | 0.261 | 0.087 | 0.306 | 0.265 |
| Claude-Sonnet4 | 0.445 | 0.647 | 0.779 | 0.188 | 0.043 | 0.111 | 0.250 |
| Gemini2.5-Flash | 0.573 | **0.941** | 0.883 | 0.442 | 0.087 | 0.639 | 0.412 |
| Gemini2.5-pro | **0.848** | 0.853 | **0.974** | **0.732** | **0.913** | **0.833** | **0.735** |
| LLama3.3-70B | 0.335 | 0.471 | 0.584 | 0.174 | 0.043 | 0.278 | 0.103 |
| Qwen3-32B | 0.384 | 0.559 | 0.545 | 0.036 | 0.261 | 0.028 | 0.132 |
| Gemma3-27B | 0.274 | 0.588 | 0.584 | 0.094 | 0.0 | 0.083 | 0.132 |
| Sarvam-M-24B | 0.201 | 0.588 | 0.143 | 0.094 | 0.0 | 0.083 | 0.132 |
| Gemma3-12B | 0.201 | 0.500 | 0.455 | 0.051 | 0.043 | 0.056 | 0.103 |

Table 11: Accuracy of different models on open-ended questions using Gemma3-27B as judge across various language understanding categories using **Telugu** prompt. *G*: Grammar, *V*: Vocabulary, *RC*: Reading Comprehension, *FK*: Factual Knowledge, *IP*: Idioms and Proverbs, *OR*: Other Reasoning.

| Category | Gemini2.5-pro | | | LLama3.3-70B | | | Qwen3-32B | | |
|---|---|---|---|---|---|---|---|---|---|
| | JS | A-1 | A-2 | JS | A-1 | A-2 | JS | A-1 | A-2 |
| Vocabulary | 0.882 | 0.867 | 0.900 | 0.412 | 0.857 | 1.0 | 0.647 | 0.591 | 0.591 |
| Linguistic Reasoning | 0.826 | 1.0 | 1.0 | 0.087 | 1.0 | 1.0 | 0.261 | 0.333 | 0.333 |

Table 12: Accuracy of responses from three LLMs as per the Gemma3-27B judge and two Human Annotators across two question categories, using **English** prompt. *JS*: Judge score. *A1, A2*: Human annotators.